# NLOMJ--NATURAL LANGUAGE OBJECT MODEL IN JAVA

Jiyou Jia
*Institute for Interdisciplinary Informatics,University of Augsburg, Germany*

Abstract: In this paper we present NLOMJ—a natural language object model in Java with English as the experiment language. This model describes the grammar elements of any permissible expression in a natural language and their complicated relations with each other with the concept "Object" in OOP (Object Oriented Programming). Directly mapped to the syntax and semantics of the natural language, it can be used in information retrieval as a linguistic method. Around the UML diagram of the NLOMJ the important classes (Sentence, Clause and Phrase) and their sub classes are introduced and their syntactic and semantic meanings are explained.

## 1. BACKGROUND

We have developed a web-based human-computer-interaction system with natural language for foreign language learning: CSIEC (Computer Simulator in Educational Communication) [1]. The kernel of this system is the natural language understanding mechanism (NLML, NLOMJ and NLDB) and the communicational response (CR). NLML(Natural Language Markup Language) is a markup language to describe the grammar of an expression in a natural language. It is produced to an expression of this natural language by a parser written according to the grammar rules and lexicon of this language [2]. We use English as the experiment language in our system. For example, the NLML for the sentence "*I come*" is

<mood>statement</mood><complexity>simple</complexity><subject><noun><type>perspronoun</type><word>I</word><numb>sing</numb><pers>first</pers><case>nom</case></noun></subject><v

erb_phrase><verb_type>verb</verb_type><tense>present</tense><numb>sing</numb><pers>first</pers
> <verb_word> come</verb_word><circum></circum></verb_phrase>

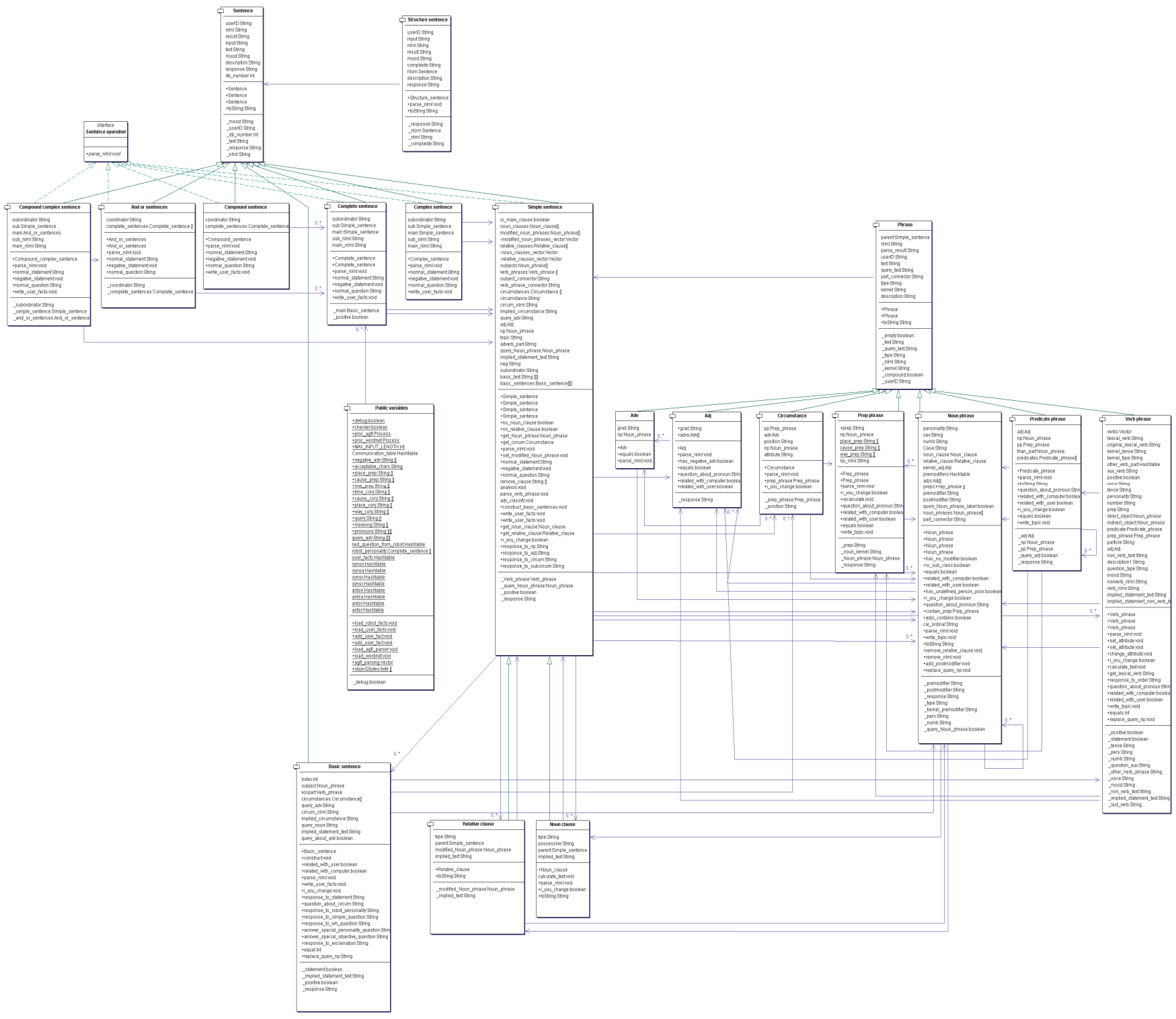

*Figure 1.* UML diagram for NLOMJ

With the simple structure of the markup language the NLML can be easily parsed by an OOP (Object Oriented Programming) language to construct the object model of the nodes in the markup language. As for the NLML the nodes are just the grammar elements of the natural language, e.g. noun phrase, verb phrase, etc. So we have selected Java, the typical OOP language, to parse the NLML and to represent the grammar elements and their relations with the concept of Object. We call this technique NLOMJ (Natural Language Object Model in Java). It enables us to extract the information we need from any permissible expression of a natural language, hence can be used as a linguistic method in intelligent information processing.

The UML diagram for the whole NLOMJ is shown in figure 1. In the following paragraphs we introduce the content of NLOMJ around this diagram in the order of sentence, clause and phrase. Limited by the paper length we can only introduce the sentence and clause more detailed.

# 2. SENTENCES

## 2.1 Super class “Sentence” and Interface “Sentence_operation”

All the sentences with diverse kinds of complexity, tense, mood and voice must have at least one subject and one verb phrase. So we use the class "Sentence" to generalize the common features of the sentences. Because there are great structure differences for the sentences with different complexity, the compound complex sentence, compound sentence, complex sentence and simple sentence are all treated as the subclass of the class “Sentence”.

In the super class Sentence there are the common attributes and operations all kinds of sentences must have. The attributes include "mood", "input", "text", "nlml", "description", “response”, etc. They are all with the type of "String". The "mood" represents the mood of the sentence therefore may have one of the values--"statement", "question", "order", "full exclamation", etc. The attribute "input" is the input text given from the user, whereas the "text" is inferred by the parser from the "input" and other attributes. So the "input" is not always the same as the "text". The “nlml” is the parsing result in the NLML format. The “description” is the syntactic and semantic description for this “text”. The attribute "response" represents this communicational response, as we want to use the NLML and the NLOMJ to form one communicational response to it.

The operations in the class “Sentence” are: get_mood() (get the "mood"), get_nlml() (get the "nlml"), get_text() (get the "text"), to_String() (get the "description"), etc.

There are two constructors for the class Sentence:

public Sentence(String user, String input, String nlml, String mood): this one is suitable for the sentences with a direct input text and a NLML, for example, for the sentences with different kinds of complexity. The “input” is given from the user and the “nlml” is the parsing production from this “input”. The “mood” is obtained from the tag “mood” in the “nlml”.

public Sentence(String user, String nlml, String mood): this is suitable for the sentences with only the NLML, but no input text, fox example, the relative clause, noun clause. These ones should be parsed as a simple sentence, and the implied text is not directly inputted, but should be calculated through the parsing.

In order to get the object model from the NLMLs a method for the parsing of the NLML should be implemented in every subclass of the “sentence”. So an Interface "Sentence_operation" is written with only this method “parse_nlml”.

## 2.2 Sentences with different complexity

The “Compound_complex_sentence” consists of a subordinate “Simple_ sentence” and an “And_or_sentence”. So we use the attribute “subordinator” with the type “String” to represent the subordinator, the “sub” with the type Simple_sentence to represent the subordinate sentence, and the “main” with the type “And_or_sentence” to represent the main sentence. But how can we semantically decompose the compound complex sentence into simple sentences? An example is:

“*If it rains today, you will not go, and I will not come*”-

The two main clauses are connected by “and” and state two possible events, which will happen with the condition contained in the subordinate clause. So the compound complex sentence with the conjunction “and” connecting N main clauses can be decomposed into N complex sentences, each of which has the same subordinate clause. As for the example above, we can decompose it into two independent complex sentences.

“*If it rains today, you will not go*” and “*If it rains today, I will not come*”.

The “And_or_sentence” consists of a coordinator (“and”, or “or”) represented by the attribute “coordinator” and several instances of the class “Complete_sentence” represented by the array “complete_sentences”.

The “Complete_sentence” consists of a subordinator represented by the attribute “subordinator”, a subordinate clause represented by the attribute “sub” with the type “Simple_sentence”, and a main sentence represented by the attribute “main” with the type “Simple_sentence”. If the subordinator of an instance of this class is null therefore the “sub” is meaningless, this instance is essentially the same as an instance of “Simple_sentence” The array “complete_sentences” in the class “Complex_compound_sentence” and the “Compound_sentence” uses this special case. But if the “subordinator” and the “sub” are all not null, this instance is the same as an instance of the class “Complex_sentence”. In other words the “Complete_ sentence” is either a “Complex_sentence” or a “Simple_sentence”. It does not exist in the real language. But this class is specially useful in describing the personality of the “chatting robot” in the “CSIEC” project, which is represented by the attribute “robot_personality” in the class “Public_ variables”, because some personalities can be stated by simple statement sentences as the facts, but others are states or behaviors happening only with some conditions, which can only be represented by the “Complex_sentence”.

The "Compound_sentence" consists of the conjunctions represented by the attribute “coordinator” and several complete sentences represented by the array "complete_sentences".

The "complex_sentence" consists of a subordinator represented by the attribute “subordinator”, a subordinate clause represented by the attribute “sub” with the type “Simple_sentence”, and a main sentence represented by the attribute “main” with the type “Simple_sentence”. The subordinator can not be null.

Summarily the simple sentence is the most elementary sentence and the end point during the parsing of all sentences with different complexities.

## 2.3 Simple sentence

This kind of sentence has at least a verb phrase. It may contain relative clause and noun clause. Because these clauses contain also verb phrases, we should deal with them at first, i.e., search the tag "noun_clause" and "relative_clause" and replace their contents with a noun phrase and a preposition phrase, respectively, then parse every noun clause and relative clause, and save them in the arrays. In the UML diagram the blue arrow with the number "0...." from the class “Simple_sentence” to the “Noun_clause” and to the “Relative_clause” indicates this relation.

We should think over the parsing with different moods. If the mood is "np", "what terse exclamation", or "about", the sentence treats essentially one noun phrase, therefore can be parsed as a noun phrase. The attribute “np” in the class “Simple_sentence” stands for this noun phrase. If the mood is "adj", or "how terse exclamation", the sentence treats essentially one adjective phrase, therefore can be parsed next as an adjective phrase. The attribute “adj” stands for this adjective phrase. If the mood is "circumstances", the sentence deals essentially with one or more circumstance phrase, therefore can be parsed next as circumstance phrases. The array “circumstances” stands for stands for these circumstance phrases. If the mood is „order“, the subject is default with the value “you”, therefore needs no further parsing. But the verb phrase and circumstances should be parsed. The array “verb_phrases” and “circumstances” stand for them, respectively. If the mood is "statement", "question", "full exclamation" or "subcircum", the sentence has at least one subject, one verb phrase and sometimes with circumstances. The array “subjects”, “verb_phrases”, and “circumstances” stand for them, respectively. Especially for the mood "subcircum" an attribute "subordinator" with the type "String" is needed.

We get the circumstances via the tag "circum" and put them in an array for them. If the sentence is a question about the circumstance, the query adverb will also be included in the array and will be specially labeled in the variable "query_adv" with the type "String".

The sentence may have several subjects. They are all represented by instance of the class “Noun_phrase” and saved in the array “subjects”. If the sentence is a question about the subject, the attribute “query_noun_phrase” equals the subject.

In order sentences and some clauses like infinitive, gerund, etc, there may be a negative phrase like "don't", "not". So it should be parsed firstly via the tag "neg". Finally we get the verb phrase via the tag "verb_phrase". If the content of the tag "verb_phrase_connector" is null, there is only one verb phrase; otherwise there are several and we get them sequentially via the tags "verb_phrase_part". The verb phrases are put into the array “verb_phrases”.

The key point is how to combine these subjects, verb phrases and circumstances. The conjunction connecting the noun parts (part_connector) and the conjunction connecting the verb phrases (verb_phrase_connector) are: “and”, “or”, “neither…nor”. It is common that either the subject or the verb phrase is made up of several parts, for example, N parts. In this case we can decompose this simple sentence into N basic sentences. If the conjunction is “and”, these N basic sentences are independent; if it is “or”, the relation among these basic sentences is single choice; if it is “neither_nor”, we can form N independent basic sentences through negative operation. The circumstances are suitable for every basic sentence.

The method in the "Simple_sentence", “contruct_basic_sentences()”, implements this operation of combination. After this operation the basic sentences are stored into the two dimensional array "basic_sentences", whose length is the length of the subject phrases times the length of the verb phrases. Additionally in this process the values of the attribute "text" and "description" are also calculated according to the “mood” and the element values of "basic_sentences".

The concept "basic sentence" we use here means the sentence with just null (for order sentence) or just one subject phrase, and one verb phrase. In the NLOMJ it is represented with the class “Basic_sentence”. As the diagram shows, it is just a subclass of the class “Sentence”, but does not implement the Interface “Sentence_operation”, because its instance is not formed by parsing the NLML, but constructed by the given subject phrase, verb phrase, circumstances, and other parameters.

In the "Basic_sentence" the "text" and "description" are calculated in the method "construct" according to the mood, the subject phrase, the verb phrase and the circumstance. We should pay attention to the calculation of "text" according to different moods.

If the mood is “statement”, there are several situations to be considered. Besides the normal simple statement sentence, the relative clause and the noun clause are also dealt with as a statement sentence. For the normal simple statement sentence, or the noun clause or relative clause whose query word is just the subject, the "text" can be obtained by the addition of the subject phrase plus the verb phrase plus the circumstance. For the noun clause or relative clause beginning with a query word which is not the subject, the phrases order is not the same as the normal simple sentence. The query phrase (either as Noun phrase or as query adverb) should be put at the beginning, then the subject, then the other parts in the verb phrase, at last the circumstances. The position of the circumstances is also important. Some circumstances must be put at the beginning of the text, some must be at the end, and some have flexible position. The position of the circumstance is decided by the value of attribute "position" in the class "Circumstance".

If the mood is order, the "text" can be obtained by the value of "neg" plus the verb phrase plus the circumstance.

If the mood is question, it is complicated. If the subject itself is a query noun clause, we can obtain the "text" by the addition of the subject with verb phrase and the circumstance; otherwise, we should put the query noun or query adverb at the beginning, then the auxiliary verb, the subject phrase, then the other part of the verb phrase, and at last the circumstance. Whether the question begins with a query adverb is decided by the value of the attribute "query_adv", which is transferred from the constructor, which is calculated by the analysis of the circumstances in the “Simple_sentence”. If the "query_adv" is null, the query word is not an adverb. Whether the subject itself is a query noun is decided by the method "get_query_text" in the Noun_phrase. If it returns a value with null, it is not query word; otherwise it is. Whether there is a query noun in the verb phrase is decided by the method "get_query_text" in the Verb_phrase. If it returns a value with null, it is not query word; otherwise it is. It there is no query word both in subject and verb phrase, the query word should be a noun phrase, which is the object of a prepositional phrase functioning as a circumstance. This can be calculated in the analysis of the circumstances in the “Simple_sentence”.

If the mood is full exclamation beginning with "what", the "text" can be obtained by the addition of "what" with the object phrase in the verb phrase, then the subject phrase, then the other parts in the verb phrase.

If the mood is full exclamation beginning with "how", the "text" can be obtained by the addition of "how" with the adjective phrase in the predicate phrase of the verb phrase, then the subject phrase, then the other parts in the verb phrase.

# 3. CLAUSES

In NLOMJ the relative clause and noun clause are defined as two subclasses of the "Simple_sentence", for they either are full simple sentences or can be extended to semantic equal simple sentences. The attributes and operations in the "Simple_sentence", such as "text", "subjects", "verb_phrases", "construct_ basic_sentences", etc, can be still applied in the two subclasses. Additionally some special attributes and operations are needed.

## 3.1 Noun clause

There are several types of noun clauses:

"that", "whether", "whether or not": In these three types the noun clause consists of a preceding word and then a simple statement sentence without noun clause(and "or not" at the end for the type "whether or not"). So the "implied_text" equals the "text" here.

"query clause": the implied text is a question and can be obtained with the same method as that used in obtaining the "text" in the Basic_sentence with the mood

"question". The key is how to get the auxiliary verb and the other parts in the verb phrase, which is realized in the class Verb_phrase. For example in the sentence "*I want to know when you will come here.*" the implied question is "*when will you come here?*"

"query_to": in this type there is no subject, so it is more difficult to infer the implied question. If the noun clause is the subject of the main sentence, this question may be suitable to all people. For example the sentence "*how to finish the work is still under discussion*." implies the question "*"How does a person finish the work?*". In this situation we assign "*a person*" as the subject of the question. If the noun clause is the object in the main sentence or the object in a prepositional phrase, we can assign the subject in the main sentence as the subject of the noun clause. e.g. "*I don't know what to do next.*" implies the question: "*what do I do next?*".

For the parsing of the noun clause the main sentence containing it plays also an important role. So there is an attribute in the class Noun_clause representing its main sentence, the "parent" with the type "Simple_sentence".

"normal_to": in this case there is also no subject in the noun clause which can be extended to a simple statement sentence. If this type of infinitive appears as the object of some special verbs, which are shown in the various construction forms of the verb phrase, we don't have to obtain the implied statement with great efforts, because the subject of the infinitive clause is the same as the subject in the main sentence and the whole verb phrase including both the main verb word and the infinitive clause expresses a mental state of the subject that the subject will do, should do, is doing, or have done something, or be in some state, and this semantic meaning can not be expressed by the main verb word or the infinitive clause alone, e.g. "*He agrees to come today*".

But if the infinitive clause follows an object clause which is the semantic subject of this clause, this whole sentence expresses the mental state of the subject that the object in this main sentence will do, should do, is doing, or have done something, or be in some state. Then this implied statement sentence can be obtained according to the verb type and the tense of the main sentence.

## 3.2 Relative clause

The relative clause is a full relative clause if it has a subject; otherwise it is a terse relative clause, whose actual subject is evidently the noun phrase it modifies. The full relative clause starts with a query word (if there is none it is "which") which points to the modified noun phrase in front of it. So we can infer the implied statement sentence by two steps: at first get the statement sentence with the query noun phrase in the way we use in the noun clause, secondly replace the query word with the modified noun phrase. The modified noun phrase is represented by the attribute "modified_noun_ phrase" which is transferred in the parsing of its modified noun phrase in the main sentence (parent) by the method "set_modified_Noun_ phrase (Noun_ phrase np)". For example the sentence "*The man whom you met yesterday is our boss.*" implies: "*you met the man yesterday.*"

For the terse sentence it is difficult to set the tense of the implied statement sentence. If it is the present participle, the implied statement sentence may have the tense "progressive". If it is the past participle, the implied statement sentence may have the tense perfect or past. This character is labeled by the tag "voice" with the value "passive". If it is passive infinitive clause, we can set the tense with the value "future" or use the model verb "should". This character is labeled by the tag "voice" with the value "passive" and "tense" with the value "future".

# 4. PHRASES

## 4.1 Super class "Phrase"

The sentence consists of various phrases, so at the end of the parsing of the sentences comes the parsing of the phrases. On other side the phrase exist in the sentence and have its syntax und semantic function in the context of its existing sentence. The super class "Phrase" in NLOMJ represents the common feathers of the phrases. Its attributes include the "nlml", "text", and "description" which are similar with those in the class "Sentence". Besides it has also the following ones: the "parent" with the type "Simple_sentence" representing the simple sentence where this phrase exists; the "type" with the type "String" representing the small category of this phrase; the "part_ connector" with the type "String" representing the conjunction word if this phrase consists of several parts; the "kernel" with the type "String" representing the kernel word in the phrase; the "query_text" with the type "String" representing the content of this phrase if it is a query word; the "parse_result" with the type "String" representing the temporal variable for the "nlml", which is set to the same value as the "nlml" at the beginning of the parsing and then changed through the process of the parsing.

The constructor for this class is: public Phrase( String nlml_result, Simple_ sentence ss). The parameter "ss" is the Simple_sentence where this Phrase exists, the "nlml_result" is the NLML this Phrase should parse.

The operations in this class are similar with those in the class "Sentence", like "get_text()", "toString()", etc. Every subclass has some special attributes and its own method "parse_nlml()" to parse the "nlml" to form an instance of this subclass. Now we introduce these subclasses.

## 4.2 Adverb

The class “Adv” represents the adverb and has two attributes: "grad" and "np". The string "grad" represents the grad of the adverb and is obtained via the tag "grad", i.e. "abso"(absolute), "comp"(comparative), or "supl" (superlative). The "np" with the type "Noun_phrase" is obtained through the parsing of the other parts in the NLML, if the type of this adverb is "so_that", "so_as", "enough_to", "too_to", or "adv_than". We do so because these types of sentences express the result or the extent of the adverb with a compared noun phrase, a statement sentence or a noun clause (infinitive clause), all of which can be parsed with the class “Noun_phrase”.

## 4.3 Adjective

The class “Adj” represents the adjective. It has also two attributes: "grad" and "advs". The "grad" is similar with that in “Adv”. The array "advs", whose elements are instances of the class “Adv”, represents the adverbs modifying this adjective.

## 4.4 Circumstance

The class “Circumstance” represents the circumstance. It can be adverb, prepositional phrase or noun clause (infinitive clause or participle clause), what is decided by the attribute “type”. Its attribute “position” represents the position of the circumstance in the sentence: pre, mid or post, and the “attribute” represents its semantic function in the whole sentence: place, time, way, and others.

## 4.5 Prepositional phrase

The class “Prep_phrase” represents the prepositional phrase. It has the attribute "prep" with the type "String" representing the preposition word and the attribute "np" with the type "Noun_phrase" representing its object.

## 4.6 Noun phrase

The class “Noun_phrase” represents the noun phrase and has many special attributes such as “personality”, “number”, “case”, “sex”, etc. It may have several parts connected by conjunction and every part can be in the normal form: pre modifiers + kernel + post modifiers. The kernel can have different types, like “countable noun”, “number”, etc, what is decided by the attribute “type”. The pre modifiers can be adjective, article, determiner, quantifier, etc. The post modifier can be prepositional phrase or relative clause. The noun clause as a noun phrase is specially considered.

## 4.7 Predicate

The class “Predicate_phrase” represents the predicate phrase. It can be adjective, noun phrase or prepositional phrase, what is decided by the attribute “type”. If it is adjective, a noun phrase or noun clause can be connected to it by “as…as”, “than”, “too…to”, “enough…to”, or “so…that” to express the compared object, the extent or the result of the adjective.

## 4.8 Verb phrase

The class “Verb_phrase” represents the verb phrase. It is the most important and complicated phrase in constructing a sentence and may consist of all other types of phrases. It describes the actions or the states of the subject, or the subject’s relation with other things or persons that can be described by direct and (or) indirect objects represented by the two instances of “Noun_phrase” (“direct_object” and “indirect_object”), or predicates represented by the instances of “Predicate_phrase” (“predicate”). The concrete pattern of the verb phrase is decided by the attribute “verb_type”. In the parsing process at first the attributes about the verb are obtained, such as “personality”, “number”, “voice”, “tense”, “kernel_tense”, “verb_type”, etc. Then the other parts in the verb phrase are obtained corresponding to the “verb_type”. For example, if the “verb_type” is “be”, the “predicate” should be obtained. At last the verb words are obtained successively.